\documentclass[final]{cvpr}

\usepackage{times}
\usepackage{epsfig}
\usepackage{graphicx}
\usepackage{amsmath}
\usepackage{amssymb}
\usepackage{multirow}
\usepackage{units}

\usepackage[pagebackref=true,breaklinks=true,colorlinks,bookmarks=false]{hyperref}

\def\cvprPaperID{****} 
\def\confYear{CVPR 2021}

\begin{document}

\title{SAIC\_Cambridge-HuPBA-FBK Submission to the EPIC-Kitchens-100 Action Recognition Challenge 2021}

\author{Swathikiran Sudhakaran$^{1}$ Adrian Bulat$^{1}$ Juan-Manuel Perez-Rua$^{1}$ Alex Falcon$^{2}$ \\[0.3cm] Sergio Escalera$^{3,4}$ Oswald Lanz$^{2}$ Brais Martinez$^{1}$ Georgios Tzimiropoulos$^{1}$ \\[.3cm] 
    $^{1}$Samsung AI Center, Cambridge, UK\\
	$^{2}$Fondazione Bruno Kessler - FBK, Trento, Italy\\
	$^{3}$Computer Vision Center, Barcelona, Spain\\
	$^{4}$Universitat de Barcelona, Barcelona, Spain\\
}

\maketitle

\begin{abstract}
   This report presents the technical details of our submission to the EPIC-Kitchens-100 Action Recognition Challenge 2021. To participate in the challenge we deployed spatio-temporal feature extraction and aggregation models we have developed recently: GSF and XViT. GSF is an efficient spatio-temporal feature extracting module that can be plugged into 2D CNNs for video action recognition. XViT is a convolution free video feature extractor based on transformer architecture. We design an ensemble of GSF and XViT model families with different backbones and pre-training to generate the prediction scores. Our submission, visible on the public leaderboard, achieved a top-1 action recognition accuracy of 44.82\%, using only RGB.
\end{abstract}

\section{Introduction} \label{sec:intr}
Video content understanding is one of the widely researched areas in computer vision with several applications ranging from automated surveillance to robotics, human computer interaction, video indexing and retrieval, \etc., to name a few. Egocentric action recognition is a particularly challenging sub-task of video content understanding. Egocentric videos are captured using wearable cameras and are often characterized by the presence of a cluttered environment containing several objects and egomotion caused by movement of the camera wearer. Recognizing the action present in a video requires extraction of fine-grained spatio-temporal features that can discriminate one action from another. EPIC-Kitchens-100~\cite{epic} is the largest egocentric action recognition dataset with ~90K video segments composed of 97 verb and 300 noun categories. The verb and noun labels of a video segment is combined to form its action label.

To participate in the challenge, we used two different video action recognition models composed of entirely different building blocks and feature aggregation strategy.
\begin{itemize}
    \item GSF\cite{gsf}: A plug and play module that can transform 2D CNNs into a high performing spatio-temporal feature extractor;
    \item XViT\cite{xvit}: A convolution free transformer based architecture for efficient video representation learning
\end{itemize}

Gate-Shift-Fuse (GSF) is a CNN based architecture that captures local relationship which introduces an inductive bias about the 3D structure of video frames within a small spatio-temporal receptive field.
On the other hand XViT, a transformer based model, captures global information and learns geometric relationship between the pixels. While GSF relies on the inductive bias owing to the locally connected convolution layers for feature extraction, XViT disregards any prior about the data and learns the relevant patterns in it that are suitable for addressing the end task. 
Thus the two models used in the challenge extract different features and are highly complementary to each other. We deployed an ensemble of the two model families to participate in the challenge. The final score is obtained by averaging the prediction scores from the individual members in the ensemble.

\begin{table*}[t]
	\centering
	\begin{tabular}{c|c|c|c|c|c}
		Method & Backbone & Pre-training & Verb & Noun & Action \\ \hline \hline
		\multicolumn{6}{c}{Validation set} \\ \hline
		\multirow{3}{*}{GSF} & IncV3 & Kinetics400 & 68.89 (90) & 51.42 (75.49) & 43.11 (64.19)
		\\ \cline{2-6}
		& Res-50 & Kinetics400 & 68.88 (90.44) & 52.73 (76.37) & 43.84 (64.95 \\ \cline{2-6}
		& Res-101 & ImageNet & 69.06 (90.33) & 53.18 (75.81) & 44.48 (64.68) \\ \hline
		XViT & ViT-B/16 & Kinetics400 & 68 (90.08) & 55.63 (78.86) & 44.91 (65.97) \\ \hline
		Ensemble & \multicolumn{2}{c|}{-} & 70.86 (91.67) & 56.7 (79.9) & 46.88 (68.18) \\ \hline \hline
		\multicolumn{6}{c}{Test set} \\ \hline
		Ensemble & \multicolumn{2}{c|}{-} & 68.16 (90.01) & 55.49 (78.98) & 44.82 (65.45) \\ \hline
	\end{tabular}\vspace{.1in}
	\caption{Performance of the models on the validation set (top) and test set (bottom) of EPIC-Kitchens 100 dataset. Ensemble score is generated by averaging the scores of individual models.}
	\label{tab:ensemble}
\end{table*}

\section{Models} \label{sec:models}

We describe details of the two model families in this section.

\subsection{GSF}

GSF, an extension of GSM~\cite{gsm}, is a light weight feature encoding module capable of converting a 2D CNN into an efficient and effective spatio-temporal feature extractor. The output features from a spatial convolution layer of the 2D backbone is first applied to a gating module, composed of a light-weight 3D convolution kernel,
to generate grouped spatial gating. The spatial gating is then applied to the input features to obtain group-gated features and residual. Forward and backward shifting in time is then applied to the group-gated features. In GSM, the time-shifted features are combined with the residual using addition operation. GSF extends this simple fusion with a data dependent weighted channel fusion mechanism using a convolution layer. The resulting spatio-temporal features are then propagated to the next layer of the backbone CNN for further processing.

\subsection{XViT}
Vision transformers~\cite{vit} can be extended for video recognition by extending the self attention mechanism between tokens within a frame to tokens from other frames as well. However, this will increase the complexity quadratically with the increase in the number of frames. To make the model tractable XViT~\cite{xvit} proposes efficient space-time mixing attention as follows.

Let $\mathbf{q}_{s,t}\in \mathbf{R}^{1\times d_h}$, $\mathbf{k}_{s,t}\in \mathbf{R}^{1\times d_h}$ and $\mathbf{v}_{s,t}\in \mathbf{R}^{1\times d_h}$ be the query, key and value at a spatial location $s$ and temporal location $t$. Then the self-attention $\mathbf{y}_{s,t}$ is computed as 
\begin{equation}
    \mathbf{y}_{s,t} = \sum_{s'=0}^{S-1} \textrm{softmax}\{(\mathbf{q}_{s,t} \cdot \tilde{\mathbf{k}}_{s',-t_w:t_w})/\sqrt{d_h}\} \tilde{\mathbf{v}}_{s',-t_w:t_w}  \label{eqn:self-attn}
\end{equation}
with
\begin{eqnarray}
\tilde{\mathbf{k}}_{s',-t_w:t_w} = [\mathbf{k}_{s',t-t_w}(d_h^{t-t_w}), \dots, \mathbf{k}_{s',t+t_w}(d_h^{t+t_w})] \\
\tilde{\mathbf{v}}_{s',-t_w:t_w} = [\mathbf{v}_{s',t-t_w}(d_h^{t-t_w}), \dots, \mathbf{v}_{s',t+t_w}(d_h^{t+t_w})]
\end{eqnarray}
where, $\mathbf{k}_{s',t'}(d_h^{t'})$ and  $\mathbf{v}_{s',t'}(d_h^{t'})$ denotes the operator for indexing the $d_h^{t'}$ channels  from $\mathbf{k}_{s',t'}$ and $\mathbf{v}_{s',t'}$, respectively.

The video transformer model used in this challenge is constructed by replacing the self-attention in \cite{vit} with Eqn.~\ref{eqn:self-attn}

\section{Experiments} \label{sec:expts}
We describe the implementation details of the two model families along with their training and testing settings in this section.
\subsection{Implementation Details} \label{sec:impl_det}
\noindent \textbf{GSF.} Gate-Shift-Fuse Networks are instantiated by plugging in GSF to the backbone layers of a 2D CNN. For the challenge, we instantiated three different models by changing the backbone CNNs. This includes InceptionV3, ResNet50 and ResNet101. The GSF variant of InceptionV3 and ResNet50 are first pre-trained on Kinetics400 dataset while for ResNet101, we used the ImageNet pretrained weights and directly trained the model on EPIC-Kitchens-100 dataset.

\noindent \textbf{XViT.} Backbone used is the base architecture ViT-B/16 from \cite{vit} with 12 transformer layers each with 12 attention heads and an embedding dimension of 768. Each frame from the video is first divided into non-overlapping patches of size 16$\times$16 and are then applied to a linear layer for vectorization. The temporal window $t_w$ is set as 1.

\noindent \textbf{Training.} We trained all our models using SGD with momentum (0.9) and a cosine scheduler with linear warmup. The base learning rate for GSF models are set as 0.01 for a batch size of 32 while XViT is trained with a base learning rate of 0.05 and a batch size of 128. GSF models are trained for 60 epochs and XViT is trained for 50 epochs. 
16 frames uniformly sampled from the input video clip are applied as input to all the models. We also applied temporal jittering during training, as done in \cite{tsn}. All models are trained in a multi-task classification setting using three classification layers to predict verb, noun and action labels. We generated the action labels by combining the verb and noun label of the video provided with the dataset to obtain a total of 3806 action categories in the training set. More details regarding training can be found in \cite{gsf} and \cite{xvit}.

\noindent \textbf{Testing.} We sample 2 clips consisting of 16 frames during testing. From each frame, 3 spatial crops are generated. Thus, from each video, we generate 6 clips. The prediction score from each of the 6 clips are averaged to obtain the video prediction.

\subsection{Results}

Tab.~\ref{tab:ensemble} lists the performance of the various models used for the challenge. The top part of the table shows the results on the validation set. From the validation set results, one can see that GSF is strong on verb prediction while XViT results in a better performance on noun prediction. This shows that GSF is a powerful model for temporal reasoning. On the other hand, the presence of global spatial receptive field of XViT enables it to perform as a strong object recognition model. Combining the prediction scores obtained from both model families improves the performance considerably, showing their complementarity in extracting spatio-temporal features. The bottom part of the table shows the performance on the test set, which is visible on the leaderboard. Note that all model developments have been done on the validation set and evaluation of individual models is not done on the test set to tune the models' performance. This shows that our ensemble generalizes well to the test data.

\section{Conclusion} \label{sec:concl}

In this report, we summarized the details of the two model families used for participating in the EPIC-Kitchens-100 Action Recognition Challenge 2021. The improved performance of the ensemble consisting of the two model families shows that the two models are complementary to each other. This resulted in achieving a top-3 action recognition performance on the leaderboard.

\section*{Acknowledgements}
FBK gratefully acknowledge the support from Amazon AWS Machine Learning Research Awards (MLRA).

{\small
\bibliographystyle{ieee_fullname}
\bibliography{epic100_report}
}

\end{document}